\documentclass[11pt]{article}
\PassOptionsToPackage{table,xcdraw}{xcolor}
\usepackage[final]{acl}

\usepackage{times}
\usepackage{latexsym}
\usepackage{tikz}
\usepackage{amssymb}
\usepackage{pifont}
\usepackage{subcaption}
\usepackage{amsmath}
\usepackage{multirow}
\usepackage{booktabs}
\usepackage{bigstrut}
\usepackage{arydshln}
\usepackage[most]{tcolorbox}
\usepackage{xcolor}
\usepackage{makecell}
\tcbuselibrary{skins}
\tcbuselibrary{breakable}
\usepackage{colortbl}
\usepackage{array}
\usepackage[T1]{fontenc}
\usepackage[utf8]{inputenc}
\usepackage{microtype}
\usepackage{graphicx}
\usepackage{listings}
\newtcblisting{promptlisting}[1]{
  breakable,
  colback=gray!4,
  colframe=gray!40,
  listing only,
  listing options={basicstyle=\ttfamily\scriptsize,breaklines=true,keepspaces=true,columns=fullflexible},
  left=4pt,right=4pt,top=1pt,bottom=1pt,
  fonttitle=\small\bfseries,
  coltitle=white,
  colbacktitle=gray!60,
  toptitle=2pt,bottomtitle=2pt,
  titlerule=0.3pt,
  before skip=5pt,
  after skip=5pt,
  title={#1},
}

\newcommand{\sys}{\textsc{DeltaMem}}
\newcommand{\eg}{\textit{e.g.}}
\newcommand{\ie}{\textit{i.e.}}

\title{\sys{}: Incremental Experience Memory for LLM Agents via Residual Trees}

\author{
 \textbf{Haoran Tan\textsuperscript{1}\footnotemark[1]\footnotemark[3]\footnotemark[4]},
 \textbf{Zeyu Zhang\textsuperscript{1}\footnotemark[1]\footnotemark[3]\footnotemark[4]},
 \textbf{Zhicheng Cao\textsuperscript{2}},
 \textbf{Rui Li\textsuperscript{1}\footnotemark[3]\footnotemark[4]},
 \textbf{Xu Chen\textsuperscript{1}\footnotemark[2]\footnotemark[3]\footnotemark[4]},
\\
\\
 \textsuperscript{1}Gaoling School of Artificial, Renmin University of China, Beijing, China,\\
 \textsuperscript{2}Duke University School of Medicine
\\
\texttt{\{tanhaoran1321,zeyuzhang,xu.chen\}@ruc.edu.cn}
}

\begin{document}
\maketitle
\footnotetext[1]{Co-first authors.}
\footnotetext[2]{Corresponding authors.}
\footnotetext[3]{Beijing Key Laboratory of Research on Large Models and Intelligent Governance}
\footnotetext[4]{Engineering Research Center of Next-Generation Intelligent Search and Recommendation, MOE}

\begin{abstract}
Large Language Model (LLM)-based agents increasingly rely on memory to learn from experiences over continual interactions. 
However, storing experiences as independent, flat units leads to substantial redundancy and retrieval conflicts, as similar episodes repeat overlapping content and subtle scene variations cause retrieved memories to offer contradictory guidance.
To address this, we introduce residual experience, positing that newly acquired experience is often an incremental variation of existing knowledge.
We propose \sys{}, a framework that organizes experience memory into two independent residual trees, one storing goal-conditioned task experience as reusable skills and another for scene-level environment knowledge. 
Each tree uses a root node for generalized base experiences and incremental delta nodes for subsequent variations, allowing related experiences to share a common foundation without duplication. For retrieval, a failure-penalized similarity scan locates the best match, reconstructing the full experience via root-to-match chain composition.
An autonomous consolidation mechanism distills high-frequency paths into new root nodes, enabling the trees to self-organize from general heuristics to specialized variants. 
Experiments across diverse interactive environments show that \sys{} consistently outperforms existing baselines. 
To facilitate future research, we release the code at \url{https://github.com/import-myself/DeltaMem}.
\end{abstract}

\section{Introduction}
Driven by the rapid progress of large language models (LLMs) in reasoning and planning \citep{wang2024survey}, autonomous agents show great promise in tackling long-horizon, complex sequential decision-making tasks across domains like web navigation, embodied manipulation, and scientific experimentation \citep{tan2026coarse}.
However, the inherent stateless nature and context limits of LLMs force agents to reason from scratch in each episode, without benefiting from past trials \citep{zhao2024expel}.
Equipping agents with memory has therefore become a central mechanism. Memory enables experience learning, allowing agents to accumulate past interactions, adapt to changing environments, and improve their behavioral policies over successive episodes \citep{zhang2025survey}.

\begin{figure}[t]
\centering
\includegraphics[width=\columnwidth]{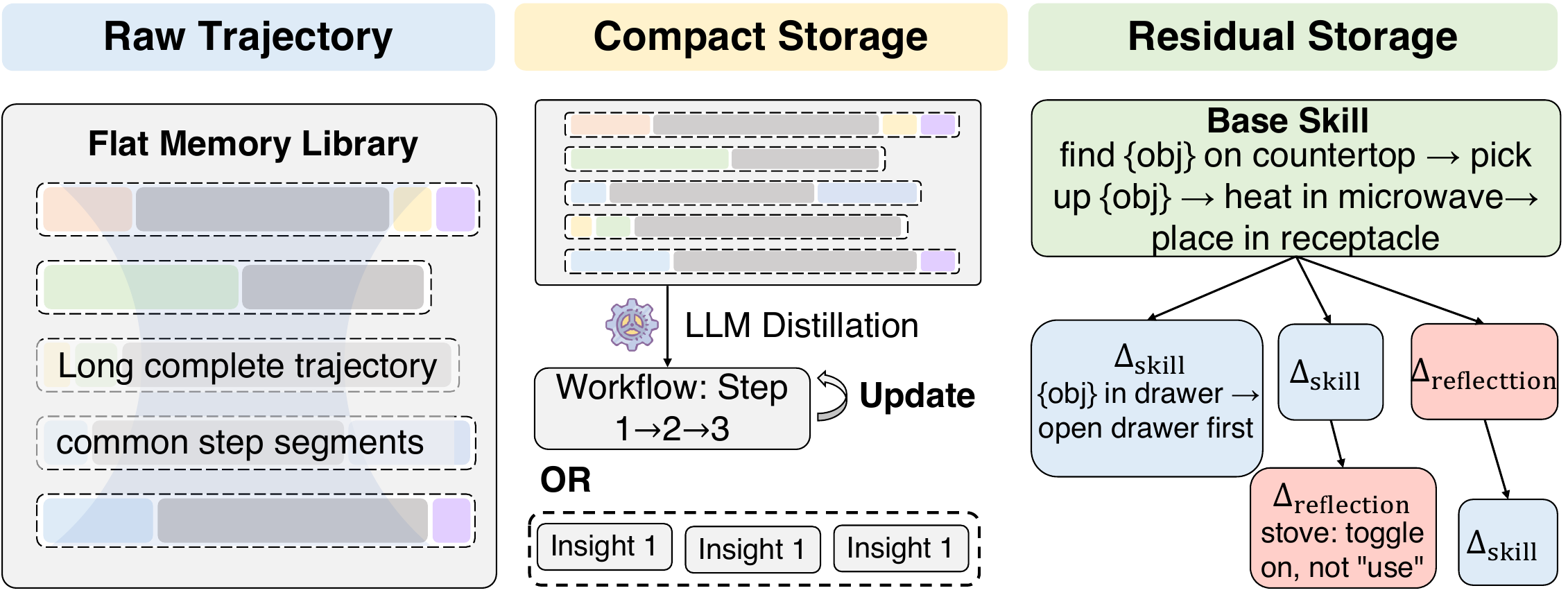}
\caption{Three memory storage paradigms. \textbf{Left}: flat trajectory storage. \textbf{Middle}: compact workflow/insight extraction. \textbf{Right}: \sys{}'s residual tree with a generalized base experience and incremental $\Delta$-nodes.}
\vspace{-0.5cm}
\label{fig:overview}
\end{figure}

Current memory mechanisms for experience learning generally fall into two broad categories, as illustrated in Figure~\ref{fig:overview}. The first category stores complete or compressed historical trajectories as memory units \citep{zheng2024synapse}. 
During inference, it retrieves the most relevant past episodes to serve as few-shot demonstrations.
The second category focuses on extracting high-level insights, reflections, or rules from past experiences rather than using raw trajectory steps \citep{ouyang2025reasoningbank, wang2024agent}. These extracted insights are then added to the agent's context as textual prompts to guide future decision-making.

Despite their wide use, both approaches treat each memory item as an \emph{independent, flat unit}, whether that item is a trajectory example or an extracted insight. 
Storing complete episodes from similar scenes creates substantial redundancy and ignores cross-episode connections. 
At read time, differences in source tasks or contexts of retrieved memories can cause conflicts, providing contradictory guidance that degrades decision quality and limits continual adaptation.
For example, two episodes of placing an object on a desk in different room layouts, when retrieved together, give conflicting advice on where to search while duplicating placement steps, mixing reusable task logic with scene-specific spatial facts in a single flat entry.

To address these limitations, we introduce \sys{}, a dual-tree residual memory framework that decouples experience into a Task-Tree for goal-conditioned action strategies and an Env-Tree for scene-level declarative knowledge. 
Rather than saving complete trajectories or extracting insights into flat and isolated repositories, each tree stores some generalized base experience from task or environment as root nodes and records new episodes as compact residual delta nodes, preserving only their incremental differences.
During retrieval, a global similarity scan with failure-node penalty reconstructs a coherent and conflict-free experience context via root-to-match chain composition.
Inspired by neuroscientific memory consolidation \citep{mcgaugh2000memory}, high-frequency convergent paths are autonomously distilled into new root nodes, allowing the memory to self-organize over time. 
To facilitate community adoption, we release the code at \url{https://github.com/import-myself/DeltaMem}. 

In summary, our contributions are as follows: \\
$\bullet$  We introduce the concept of residual experience via incremental delta storage to eliminate memory redundancy and retrieval conflicts. Based on this foundation, we propose \sys{}, a dual-tree framework that further decouples task policy from environment knowledge, enabling independent retrieval and evolution of each dimension.\\
$\bullet$ We design a tree-internal memory consolidation mechanism that distills high-frequency convergent paths into new root nodes, enabling the memory hierarchy to self-organize autonomously over successive episodes.\\
$\bullet$  We systematically evaluate \sys{} across diverse interactive environments demonstrating consistent and significant improvements over existing experience memory baselines.

\section{Related Work}

\begin{figure*}[t]
  \centering
  \includegraphics[width=\textwidth]{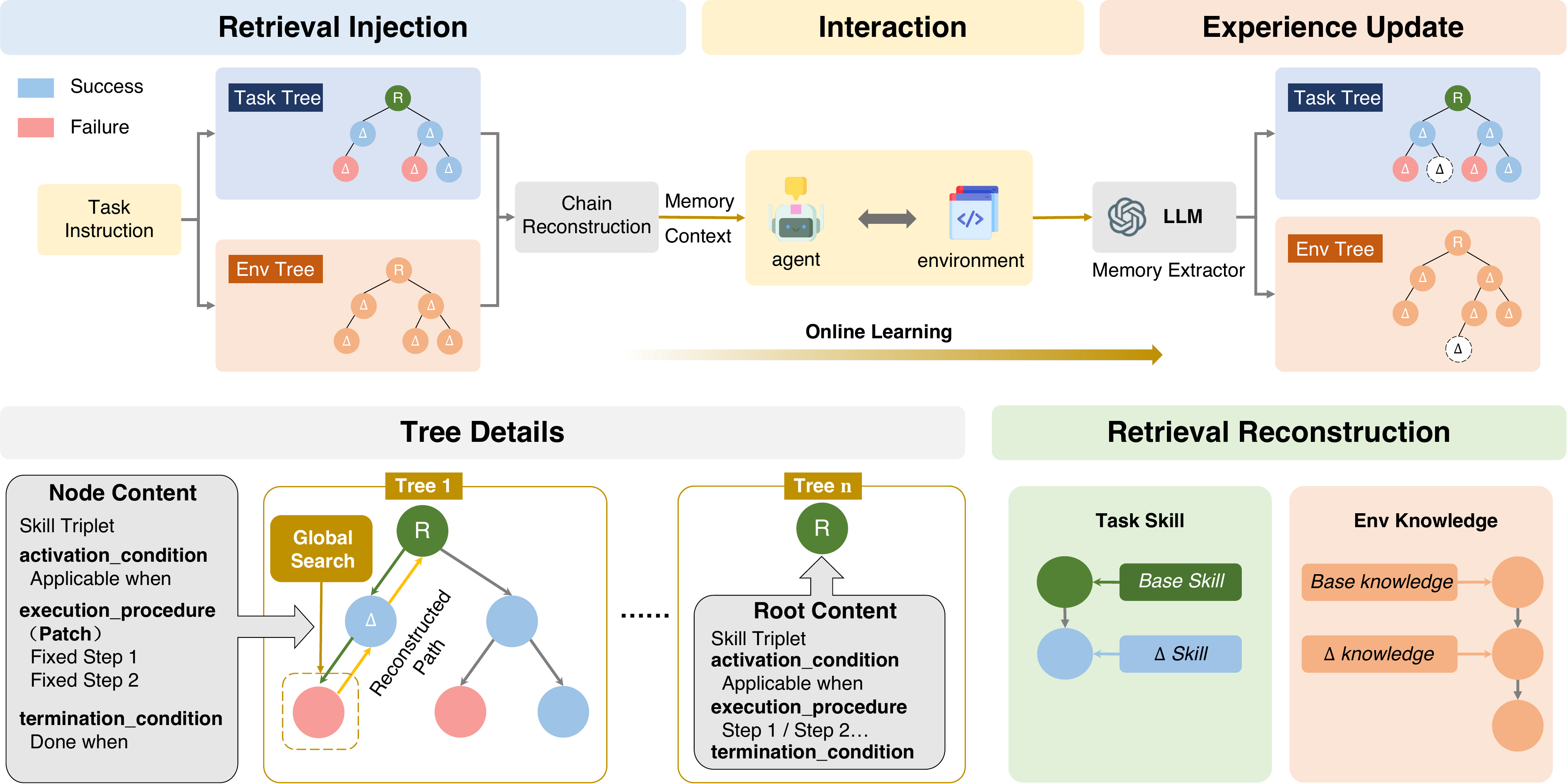}
  \caption{
    Overview of \sys{}.
    \textbf{Top}: End-to-end pipeline. A task instruction triggers retrieval from the dual trees, the reconstructed memory context is injected into the agent for environment interaction, and the resulting experience is extracted back into the trees via online learning.
    \textbf{Bottom-left}: Tree details showing node content structure, global search across all trees, and the reconstructed root-to-match path.
    \textbf{Bottom-right}: Chain reconstruction that assembles a Base Skill with $\Delta$\,Skill deltas and Base Knowledge with $\Delta$\,Knowledge deltas into a complete, conflict-free memory context for the downstream agent.
  }
  \vspace{-0.5cm}
  \label{fig:architecture}
\end{figure*}

\paragraph{LLM-Based Agents.}
LLM-based autonomous agents are commonly structured around four core components, namely \emph{profile}, \emph{planning}, \emph{memory}, and \emph{action} \citep{wang2024survey,sumers2023cognitive}.
ReAct \citep{yao2022react} establishes a Thought-Action-Observation loop grounded in environment feedback; Reflexion \citep{shinn2023reflexion} adds verbal self-criticism for within-episode refinement.
Chain-of-thought prompting \citep{wei2022chain} further improves multi-step reasoning within a single episode.
Planning has been extended via tree-search \citep{yao2023tree} and hierarchical frameworks \citep{tan2026coarse}.
Despite strong per-episode performance, these systems treat each episode independently, and cross-episode experience accumulation remains an open and active research challenge.

\paragraph{Agent Memory.}
Memory is widely studied in dialogue and personal-assistant settings \citep{zhang2025survey,tan2025membench,zhang2026memsim}.
For explicit memory, MemGPT \citep{packer2023memgpt} introduces OS-style paging to extend LLM context, Mem0 \citep{chhikara2025mem0} supports scalable long-term personal memory, Generative Agents \citep{park2023generative} maintain a perception-reflection-retrieval stream, and MemTree \citep{rezazadeh2025isolated} organizes memories into a tree structure for structured retrieval.
\citet{zhang2026explicit} further explore implicit memory by encoding past interactions into model parameters.
For interactive task agents, memory is used to accumulate executable experience across episodes.
Synapse \citep{zheng2024synapse} retrieves the most similar successful trajectory as a few-shot exemplar, but storage grows linearly; AWM \citep{wang2024agent} and ExpeL \citep{zhao2024expel} extract workflow rules into flat per-task files; ReasoningBank \citep{ouyang2025reasoningbank} tags distilled lessons as success or failure entries.
Voyager \citep{wang2023voyager} builds a growing skill library through open-ended exploration, while ReMEmbR \citep{anwar2025remembr} structures episodic memory for robot navigation.
These flat structures cannot generalize across scene variants or decouple task policy from environment knowledge.

In contrast, \sys{} addresses these limitations with a dual residual tree that decouples task policy (Task-Tree) from environment knowledge (Env-Tree) and stores each new experience as a compact delta relative to the most similar existing chain, enabling cross-scene generalization without incurring linear storage growth.
\section{Method}

\sys{} is built around two complementary principles: \emph{structural decoupling} and \emph{residual compression}.
Rather than treating each episode as an independent, flat memory unit, \sys{} separates experience into two orthogonal dimensions: goal-conditioned procedural experience and scene-level environment knowledge, storing each new interaction as an incremental delta relative to the most similar prior experience.
Figure~\ref{fig:architecture} gives a system-level overview, showing the dual-tree structure in the left panel, the global flat retrieval with failure-node penalty in the middle panel, and the root-to-match chain reconstruction that assembles a complete, noise-free experience context for the LLM in the right panel of Figure~\ref{fig:architecture}.

\begin{figure*}[t]
  \centering
  \includegraphics[width=\textwidth]{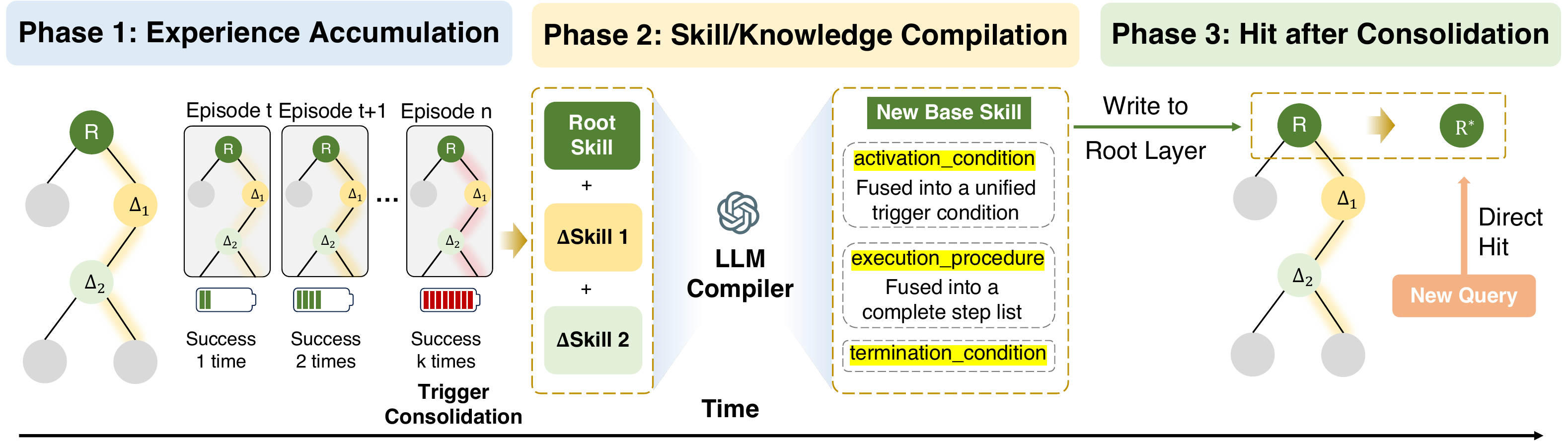}
  \caption{Autonomous memory consolidation in \sys{}. Residual nodes accumulate with success-hit counts (Phase~1); upon reaching $K_{\text{cons}}$, an LLM fuses the chain into a new root node $R^*$ (Phase~2); subsequent queries match $R^*$ directly, bypassing chain traversal (Phase~3).}
  \vspace{-0.5cm}
  \label{fig:extraction}
\end{figure*}

\subsection{Dual-Tree Memory Architecture}

Agent experience spans two qualitatively distinct dimensions. \emph{Task policy} captures goal-conditioned procedural actions specific to a given objective, whereas \emph{environment knowledge} covers scene-level declarative facts such as object features, spatial layouts, and interaction rules that remain globally consistent and task-agnostic. Conflating these two dimensions forces task-specific heuristics to mix with general scene facts, which leads to cross-task knowledge interference when the agent attempts novel goals within familiar environments.
For instance, the step-by-step action sequence for finding and placing an object belongs to task policy, while observations such as where objects of a given type tend to be stored constitute scene knowledge that transfers across many different task goals.

To achieve this decoupling, \sys{} maintains two independent hierarchical memory structures. The \textbf{Task-Tree} ($\mathcal{T}_{\text{task}}$) stores procedural experience entries, each organized as a reusable \emph{skill} with three components:
\begin{equation}
  \text{Skill} = \langle \alpha, \pi, \omega \rangle
\end{equation}
where $\alpha$ is the activation condition that specifies when this skill applies; $\pi$ is the execution procedure, a self-contained ordered action sequence; and $\omega$ is the termination condition indicating when the skill is complete. The task goal extracted from the current instruction serves as the query to the Task-Tree at inference time.

Concurrently, the \textbf{Env-Tree} ($\mathcal{T}_{\text{env}}$) stores \emph{scene knowledge entries}, each capturing environment-level declarative facts:
\begin{equation}
  \text{Knowledge} = \langle \alpha, \kappa \rangle
\end{equation}
where $\alpha$ is the scene-level descriptive trigger used for matching, and $\kappa$ represents the declarative content, such as object layouts and interaction rules. Rather than relying on a separately generated environment description, the Env-Tree query is obtained by extracting the environment relevant portion from the task instruction or initial observation (\eg{}, scene descriptions, and spatial context), decoupling it from the goal specification.
These two trees evolve asynchronously and independently; their respective retrieved chains are fused only at inference time, ensuring clean, interference-free representations for agent memory.

\subsection{Residual Tree Structure}

To directly address the massive memory redundancy caused by storing complete, flat trajectories, \sys{} organizes experiences into a residual tree architecture. Each node $v$ in either tree is formally defined as:
\begin{equation}
  v = \langle \text{type}, \text{label}, d(v), c(v), \mathbf{e}_v, P \rangle
\end{equation}
where $\text{type} \in \{\textsc{Root}, \textsc{Residual}\}$; $\text{label} \in \{\textsc{Success}, \textsc{Failure}\}$ records the episode outcome; $d(v)$ denotes the node depth; $c(v)$ is the success-hit count, incremented when successful episodes terminate at $v$; $\mathbf{e}_v$ is the dense embedding of the node's descriptive trigger; and $P$ is the actual payload, either Skill or Knowledge.

\textbf{Root nodes} serve as the foundational anchor, storing a complete, generalized base experience for a specific scene class.
\textbf{Residual nodes}, by contrast, store only a delta payload $\Delta P$. This delta encodes the incremental modification relative to its parent chain, covering additions such as extra steps, boundary-condition extensions, or newly discovered object states, without repeating shared historical content. As a result, successive interactions within variants of the same scene share a common root and record only their incremental behavioral differences, eliminating trajectory duplication.

\textbf{Write Decision.} After each episode, the system evaluates the retrieval outcome to determine the nature of the newly extracted node $v_{\text{new}}$:
\begin{equation}
  \text{type}(v_{\text{new}}) = \begin{cases}
    \textsc{Root}     & s(v^*) < \tau_{\text{base}} \\
    \textsc{Residual} & s(v^*) \geq \tau_{\text{base}}
  \end{cases}
\end{equation}
A \emph{cold start} indicates that no similar experience exists; the system inserts a self-contained root node. A \emph{warm start} appends an incremental residual delta to the best-matching retrieved chain.

\textbf{Attachment Rule.} To prevent unbounded chain growth and maintain efficient retrieval depth, the new node is mounted according to the following rule, where $r_0$ denotes the virtual root:
\begin{equation}
  \text{parent}(v_{\text{new}}) = \begin{cases}
    r_0                & \text{type} = \textsc{Root} \\
    v^*                & d(v^*) < D_{\max} \\
    \text{parent}(v^*) & \text{otherwise}
  \end{cases}
\end{equation}
If the best-matching node $v^*$ has already reached the maximum allowed depth $D_{\max}$, the new node is attached as a sibling of $v^*$ rather than further deepening the existing branch.

\subsection{Retrieval and Chain Reconstruction}

Standard flat retrieval often aggregates memories from disparate past scenarios, surfacing contradictory plans. \sys{} addresses this through a global scan combined with hierarchical reconstruction to ensure a coherent, conflict-free context.

Given a query text $q$, \sys{} scans all nodes $v$ across the corresponding tree and computes an adjusted similarity score:
\begin{equation}
  s(v) = \cos(\mathbf{e}_q, \mathbf{e}_v) - \varepsilon \cdot \mathbb{I}(\text{label}_v = \textsc{Failure})
\end{equation}
where $\mathbb{I}(\cdot)$ is the indicator function and $\varepsilon$ is a configurable failure-node penalty (default $\varepsilon{=}0.05$). The globally highest-scoring node $v^* = \arg\max_v s(v)$ is accepted as $v_{\text{match}}$ strictly if $s(v^*) \geq \tau_{\text{base}}$.

The failure penalty suppresses previously unsuccessful strategies without discarding them, turning past mistakes into retrieval guardrails, while the global flat search directly locates the best-matching node at any depth, bypassing irrelevant branches. 

Once $v_{\text{match}}$ is identified, the system assembles the root-to-$v_{\text{match}}$ path into a complete experience context:
\begin{equation}
  \mathcal{C} = \langle P^{(0)}_{v_0}, \Delta P_{v_1}, \ldots, \Delta P_{v_{\text{match}}} \rangle
\end{equation}
The reconstructed task chain $\mathcal{C}_{\text{task}}$ and environment chain $\mathcal{C}_{\text{env}}$ are then concatenated. This fusion provides the LLM with a focused, noise-free decision context that reflects the specifics of the current task.

\subsection{Experience Extraction Protocol}

Following each episode, \sys{} prompts an LLM to distill the raw trajectory, alongside the retrieved memory chain if applicable, into a structured node. We design explicit extraction templates based on the node type and episode outcome.

For the Task-Tree,
\textbf{Root-Success} extracts a complete base skill $\langle \alpha, \pi, \omega \rangle$, abstracting away environment-specific constants to maximize generalizability.
\textbf{Root-Failure} records the observed actions and unexpected environment responses without hallucinating arbitrary replacements.
\textbf{Residual-Success} extracts a minimal $\Delta\text{Skill}$ that covers \emph{only} the procedural variations not captured by the retrieved parent chain. If the parent chain perfectly covers the new episode, no redundant node is written.
\textbf{Residual-Failure} documents the exact breakdown point and the corresponding environment feedback relative to the retrieved chain.

For the Env-Tree,
\textbf{Root-Knowledge} is triggered upon the first encounter with a scene class, extracting a foundational entry $\langle \alpha, \kappa \rangle$ that maps object layouts and baseline interaction rules.
\textbf{Residual-Knowledge} is triggered during subsequent interactions with scene variants, extracting a delta $\Delta\kappa$ that documents only novel spatial relationships or newly observed rule deviations.

\subsection{Autonomous Memory Consolidation}

To prevent the residual trees from fragmenting over prolonged continual learning, \sys{} introduces an autonomous memory consolidation mechanism, illustrated in Figure~\ref{fig:extraction}. The system tracks a success-hit count $c(v)$ for each non-root node.

When a specific path proves highly reliable ($c(v) \geq K_{\text{cons}}$), the LLM is invoked to fuse the root-to-$v$ chain into a single optimized node $P^*$:
\begin{equation}
  \mathcal{C} = \langle P^{(0)}_{v_0}, \Delta P_{v_1}, \ldots, \Delta P_{v_k} \rangle \xrightarrow{\mathrm{Distill}} P^*
\end{equation}
This distilled experience $P^*$ is written back into the tree as a new root node, while the original node $v$ is marked as consolidated to prevent redundant re-triggering of the same path.

This mechanism affects both trees in distinct ways. The Task-Tree consolidates convergent, multi-step action strategies into self-contained skills. The Env-Tree solidifies stable observations into reusable scene descriptions. Over time, high-frequency interaction paths are promoted to root nodes, leaving low-frequency or highly specific variants as fine-grained residuals. The memory architecture thereby self-organizes from general heuristics into a hierarchy of specialized variants, without relying on external routing caches.

\section{Experiments}

\subsection{Experimental Settings}
\label{sec:exp}

\paragraph{Benchmarks.}

We evaluate on three interactive decision-making benchmarks spanning embodied environments and web navigation.
\textbf{ALFWorld} \citep{shridhar2020alfworld} is a text-based household environment with six task categories such as pick-and-place, heating or cooling objects, and examining items under a lamp. We evaluate on two splits: \textit{Seen} (\texttt{eval\_in\_distribution}, 140 episodes) and \textit{Unseen} (\texttt{eval\_out\_of\_distribution}, 134 episodes), each capped at 30 steps. The two splits are treated as independent evaluation runs, each starting from an empty memory bank. Since ALFWorld assigns binary reward per episode, its average reward is numerically equivalent to task success rate.
\textbf{ScienceWorld} \citep{wang2022scienceworld} is a text-based science experiment environment covering 24 task types such as measuring melting points, testing conductivity, and species classification. We evaluate on \textit{Seen} (\texttt{dev}, 194 episodes) and \textit{Unseen} (\texttt{test}, 211 episodes), also treated as independent runs from empty memory; steps per episode are task-dependent.
\textbf{WebShop} \citep{yao2022webshop} is a simulated e-commerce environment where agents must search for and purchase products matching a natural language instruction. Following the settings of previous work, we evaluate on a set of 200 episodes\citep{song2024trial}.

\paragraph{Baselines.}
We compare \sys{} with four methods, all using the same underlying LLM and prompt framework. \textbf{No Memory}, which solves each episode independently with only the system prompt and one ICL demonstration, serving as the zero-shot baseline. \textbf{Synapse} \citep{zheng2024synapse}, which retrieves the most similar successful trajectory from a dense-embedding indexed pool as a few-shot exemplar. \textbf{AWM} \citep{wang2024agent}, which induces reusable workflow descriptions from successful trajectories and maintains per-task-type workflow files updated online. And \textbf{ReasoningBank} \citep{ouyang2025reasoningbank}, which accumulates distilled lessons tagged as success strategies or failure warnings and injects the most relevant entries into the agent's decision prompt.

\paragraph{Implementation details.}
All methods query the DeepSeek-V4-flash API \citep{deepseekai2026deepseekv4} at a decoding temperature of 0 using the ReAct framework \citep{yao2022react}.
Node embeddings are computed with E5-base-v2 \citep{wang2022text}, running locally on the scenario description text.
Each benchmark retains one human-annotated task example as the in-context demonstration ($n_{\text{icl}}=1$).
Since \sys{} relies on dense-embedding similarity, the thresholds $\tau_{\text{base}}$ and $K_{\text{cons}}$ of per tree cannot be set heuristically as their appropriate values depend on each benchmark's embedding space.
We select these parameters by sampling a small subset of episodes from each benchmark and examining the empirical similarity score distribution to determine appropriate threshold ranges.
The final parameter values for each benchmark are listed in Appendix~\ref{app:params}.
Further sensitivity analysis is reported in Section~\ref{sec:sensitivity}.

\paragraph{Evaluation protocol.}
All methods follow an \emph{online learning} protocol: the agent starts from an empty memory and processes episodes sequentially, with experience extracted and memory updated after each episode. The primary metric across all benchmarks is \textbf{average reward} (AvgRew).

\begin{table}[t]
  \centering
  \small
  \setlength{\tabcolsep}{3pt}
  \caption{
    Main results across all benchmarks, measured by AvgRew\,($\uparrow$).  ALF.\,=\,ALFWorld, Sci.\,=\,ScienceWorld, Web.\,=\,WebShop.
    Best result per column is shown in \textbf{bold}.
  }
  \label{tab:main}
  \begin{tabular}{l cc cc c}
    \toprule
    \multirow{2}{*}{\textbf{Method}}
      & \multicolumn{2}{c}{\textbf{ALF}}
      & \multicolumn{2}{c}{\textbf{Sci}}
      & \textbf{Web} \\
    \cmidrule(lr){2-3}\cmidrule(lr){4-5}\cmidrule(lr){6-6}
      & \textit{Seen} & \textit{Uns.}
      & \textit{Seen}  & \textit{Uns.}
      & ---\\
    \midrule
    No Memory         & 0.7071 & 0.7985 & 0.7716 & 0.7186 & 0.3674 \\
    Synapse           & 0.7929 & 0.7015 & 0.8452 & 0.8558 & 0.5890 \\
    AWM               & 0.6429 & 0.8134 & 0.8052 & 0.7453 & 0.3337 \\
    RBank             & 0.7429 & 0.7463 & 0.7942 & 0.6898 & 0.3498 \\
    \midrule
    \textbf{\sys{}}   & \textbf{0.8500} & \textbf{0.8358} & \textbf{0.8498} & \textbf{0.8688} & \textbf{0.6254}  \\
    \bottomrule
  \end{tabular}
  \vspace{-0.5cm}
\end{table}

\begin{table}[t]
  \centering
  \small
  \setlength{\tabcolsep}{4pt}
  \caption{
    Offline memory bank utilization.
    Memory is built on the training split, then transferred to the Seen splits.
    F\,=\,Frozen (bank fixed), U\,=\,Unfrozen (bank continues updating).
    RBank\,=\,ReasoningBank.
    All values are AvgRew\,($\uparrow$). \textbf{Bold}\,=\,best per column.
  }
  \label{tab:transfer}
  \begin{tabular}{l cc}
    \toprule
    \textbf{Method}
      & \textbf{ALFWorld}
      & \textbf{ScienceWorld} \\
    \midrule
    AWM (F)               & 0.7071 & 0.8319 \\
    AWM (U)               & 0.6714 & 0.8199 \\
    \midrule
    RBank (F)             & 0.7429 & 0.7358 \\
    RBank (U)             & 0.7357 & 0.7495 \\
    \midrule
    \textbf{\sys{}} (F)   & \textbf{0.7929} & 0.8284 \\
    \textbf{\sys{}} (U)   & \textbf{0.7929} & \textbf{0.8373} \\
    \bottomrule
  \end{tabular}
  \vspace{-0.8cm}
\end{table}

\begin{table}[t]
  \vspace{0.1cm}
  \centering
  \small
  \setlength{\tabcolsep}{5pt}
  \caption{
    Component ablation on the Seen split.
    \textit{Task Only}: Task Tree only.
    \textit{Env Only}: Env Tree only.
    All values are AvgRew\,($\uparrow$). \textbf{Bold}\,=\,best per column.
  }
  \label{tab:ablation}
  \begin{tabular}{l ccc}
    \toprule
    \textbf{Method}
      & \textbf{ALFWorld}        & \textbf{SciWorld}        & \textbf{WebShop} \\
    \midrule
    \textbf{\sys{}}  & \textbf{0.8500} & \textbf{0.8498} & \textbf{0.5889} \\
    Env Tree Only    & 0.7857          & 0.7540          & 0.5112          \\
    Task Tree Only   & 0.8143          & 0.7669          & 0.5346          \\
    \bottomrule
  \end{tabular}
  \vspace{-0.5cm}
\end{table}

\begin{figure*}[!t]
  \centering
  \includegraphics[width=\textwidth]{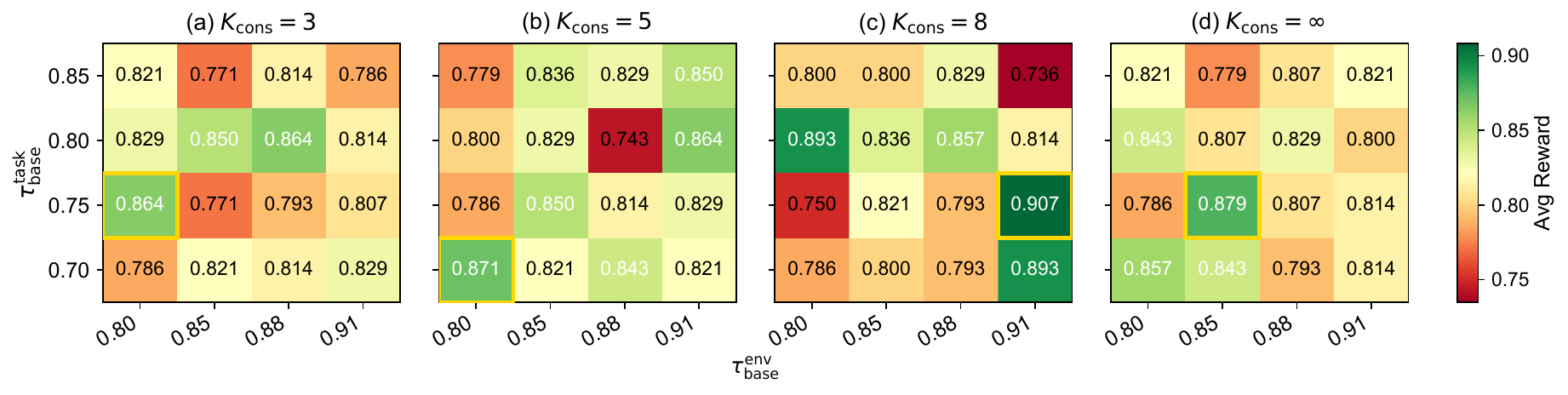}\\[-2pt]
  \includegraphics[width=\textwidth]{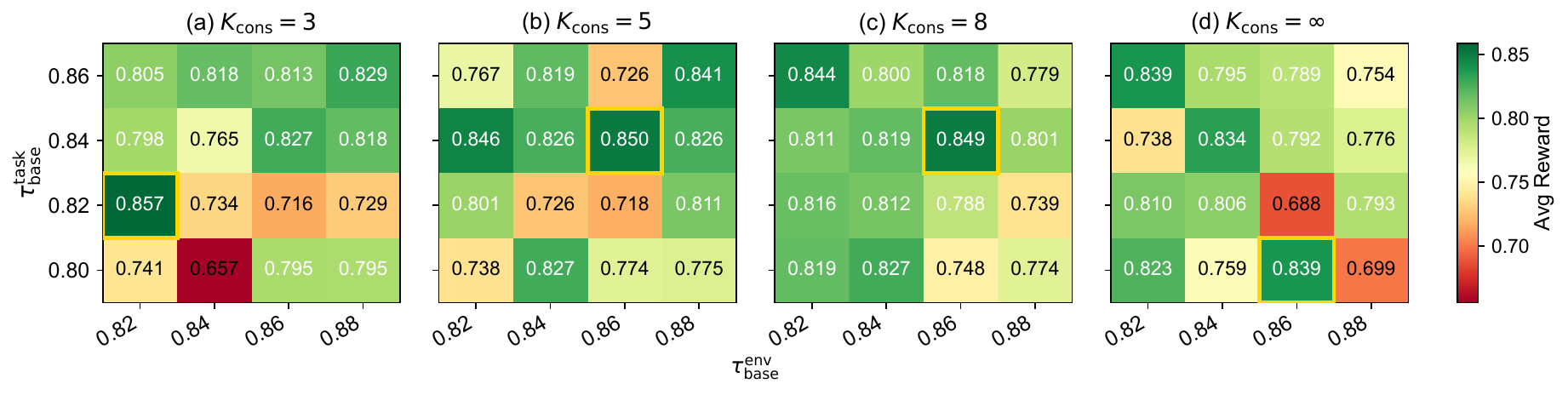}
  \caption{
    Performance across a grid of $(\tau^{\text{task}}_{\text{base}},\,\tau^{\text{env}}_{\text{base}})$ under four $K_{\text{cons}}$ levels for ALFWorld (top) and SciWorld (bottom).
    Gold box marks the per-subplot optimum for each consolidation setting.
  }
  \vspace{-0.5cm}
  \label{fig:heatmaps}
\end{figure*}

\subsection{Main Results}
\label{sec:main}

Table~\ref{tab:main} reports the main results across all benchmarks.
\sys{} achieves the best or competitive performance across all splits and benchmarks.

On ALFWorld, \sys{} consistently leads on both splits.
Synapse performs well on Seen but degrades sharply on Unseen, even falling below No-Memory, suggesting that complete-trajectory storage memorizes specific layouts without capturing transferable knowledge.
AWM struggles on Seen, likely because high-level workflow abstraction discards fine-grained action detail.
\sys{}'s residual structure generalizes naturally by abstracting a shared base and recording only incremental behavioral differences, avoiding the redundancy of storing complete trajectories.

On ScienceWorld, \sys{} achieves the best results on both splits.
The gap is more pronounced on Unseen, suggesting that the dual-tree structure accumulates increasingly reliable knowledge, while flat-storage methods plateau as memory redundancy steadily accumulates.

On WebShop, \sys{} performs on par with the best baseline.
The relatively homogeneous nature of product-search tasks limits the additional signal provided by the Env-Tree, explaining the smaller differentiation compared to the physically grounded benchmarks above.

On average, residual nodes store roughly half the tokens of root nodes, confirming that shared content is not duplicated across episodes.
Despite the structured skill format, per-memory token counts remain comparable to insight-based methods and well below full-trajectory storage; detailed statistics are in Appendix~\ref{app:token_stats}.

\subsection{Offline Memory Bank Utilization}
\label{sec:t2t}

The main experiments adopt an \emph{online} protocol in which memory is accumulated from scratch throughout evaluation.
To assess the practical value of a pre-built, reusable memory bank, we introduce a \emph{train-to-test} protocol.
First, in the \textbf{Build} phase, the agent runs on the training split under the standard online-learning protocol to accumulate a full memory bank.
Then, in the \textbf{Evaluate} phase, the pre-built bank is loaded onto the Seen splits under two settings:
\emph{Frozen}, where the bank is fixed and no updates are made during evaluation;
and \emph{Unfrozen}, where the bank serves as a warm start and continues updating online during evaluation.
Table~\ref{tab:transfer} reports results for both settings.

\sys{} achieves the best results across both benchmarks and both evaluation settings (Table~\ref{tab:transfer}), demonstrating that the pre-built residual tree transfers effectively to new episodes without re-construction from scratch.
Notably, the Frozen and Unfrozen variants of \sys{} perform comparably on ALFWorld, indicating that the memory bank is already sufficient to guide new episodes; on ScienceWorld, the Unfrozen variant yields a slight gain, consistent with continued refinement in a richer scientific environment.

In contrast, AWM's Unfrozen variant degrades on ALFWorld, suggesting that online updates to flat workflows introduce conflicting modifications; ReasoningBank shows a similar pattern.
\sys{}'s non-destructive tree structure avoids this by appending new delta nodes rather than overwriting previously stored entries.

\begin{figure*}[!t]
  \centering
  \includegraphics[width=\textwidth]{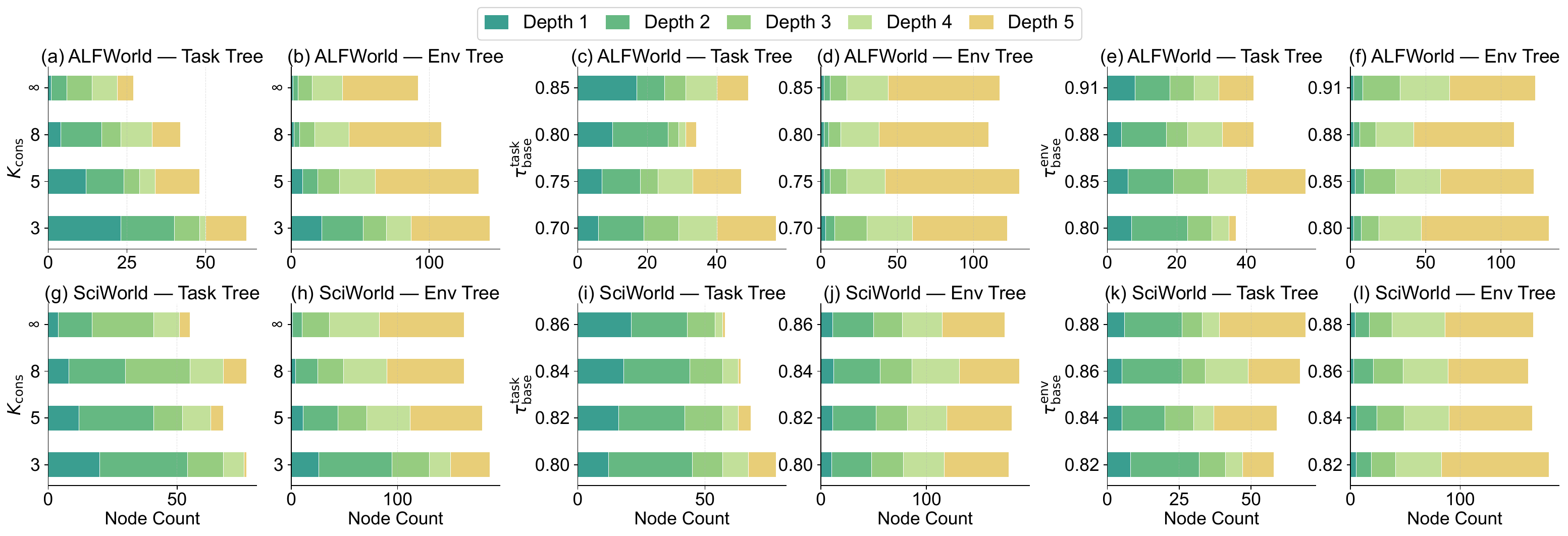}
  \caption{Node distribution by tree depth under different parameter settings. Top row: ALFWorld; bottom row: SciWorld. Stacked bars show node counts at each depth level. \textbf{Left}: varying $K_{\text{cons}}$ with thresholds fixed. \textbf{Middle}: varying $\tau^{\text{task}}_{\text{base}}$ with $\tau^{\text{env}}_{\text{base}}$ and $K_{\text{cons}}$ fixed. \textbf{Right}: varying $\tau^{\text{env}}_{\text{base}}$ with $\tau^{\text{task}}_{\text{base}}$ and $K_{\text{cons}}$ fixed.}
  \label{fig:tree_all}
  \vspace{-0.5cm}
\end{figure*}

\subsection{Ablation Study}
\label{sec:ablation}

We ablate the two memory components of \sys{}: the Task-Tree and the Env-Tree.
Table~\ref{tab:ablation} reports performance when each component is disabled, confirming that both trees provide independent and complementary contributions.

Removing either tree leads to consistent performance drops across all benchmarks.
On ALFWorld, the Task-Tree contributes more, reflecting the procedural nature of household manipulation where fine-grained action sequences are the primary bottleneck.
On ScienceWorld, both single-tree variants fall substantially below the full model, indicating that precise scene-level knowledge and goal-conditioned task policy are jointly necessary for effective scientific experiment planning.
The consistent gap between any single-tree variant and the full \sys{} confirms that the two trees capture orthogonal aspects of experience, with their combination yielding gains that neither achieves alone.
On WebShop, both single-tree variants also show clear drops, confirming that the dual-tree design benefits even relatively homogeneous task distributions such as product search.

\subsection{Parameter Sensitivity Analysis}
\label{sec:sensitivity}

We study how the retrieval thresholds $\tau^{\text{task}}_{\text{base}}$, $\tau^{\text{env}}_{\text{base}}$, and the consolidation threshold $K_{\text{cons}}$ jointly affect task performance and overall tree topology.

\paragraph{Performance sensitivity to retrieval thresholds.}
Figure~\ref{fig:heatmaps} shows performance across all threshold combinations under four consolidation settings.
When consolidation is active, \ie{} $K_{\text{cons}}<\infty$, the performance landscape is broad and flat: a wide range of threshold pairs yields near-optimal results, confirming that \sys{} is not sensitive to the exact threshold values.
As $K_{\text{cons}}$ increases toward $\infty$ (no consolidation), the optimal region narrows and the gap between the best and worst threshold configurations widens.
This suggests that consolidation has a regularizing effect on retrieval quality, making the system more tolerant of suboptimal retrieval threshold settings.

\paragraph{Effect of $K_{\text{cons}}$ on tree structure.}
As shown in Figure~\ref{fig:tree_all} (left), the tree depth distribution shifts consistently with $K_{\text{cons}}$ across both benchmarks.
When $K_{\text{cons}}$ is small, high-frequency paths are consolidated into root-level nodes more often, so Depth-1 nodes account for a larger share of the tree.
As $K_{\text{cons}}$ increases toward $\infty$, consolidation rarely triggers and nodes accumulate at greater depths instead.
This is especially clear in the Env-Tree, where Depth-1 count drops monotonically as $K_{\text{cons}}$ increases, confirming that consolidation effectively compresses convergent scene observations into reusable root entries.

\paragraph{Effect of retrieval thresholds on tree structure.}
Figure~\ref{fig:tree_all} (middle and right) vary one threshold at a time while keeping the other fixed.
A stricter task threshold $\tau^{\text{task}}_{\text{base}}$ leads to more root-level entries in the Task-Tree: when an episode fails to match any existing node above the threshold, it is stored as a new root rather than a residual delta.
A looser threshold, by contrast, attaches more episodes to existing chains and produces deeper trees.
The Env-Tree shows the same pattern under $\tau^{\text{env}}_{\text{base}}$, though the effect is somewhat weaker given the more constrained vocabulary of scene descriptions.
In both cases the performance landscape remains stable across the tested range (Figure~\ref{fig:heatmaps}), which is consistent with the threshold selection strategy described in detail in Section~\ref{sec:exp}.

\section{Conclusion}

We presented \sys{}, a dual-tree residual memory framework designed to resolve the redundancy and retrieval conflicts of existing flat-storage approaches. By decoupling task policies from environment knowledge and organizing experiences as residual delta nodes, \sys{} compresses storage while preserving full semantic content for cleaner reuse. Through failure-penalized global retrieval and chain reconstruction, the framework successfully provides agents with a coherent and conflict-free context. Additionally, autonomous memory consolidation promotes high-frequency paths into a self-organizing hierarchy of base skills and specialized variants. Experiments across three benchmarks demonstrate consistent improvements in both task success rates and storage efficiency, offering a robust foundation for scalable and continually-learning agents.

\section*{Limitations}

\sys{} relies on LLM-generated experience extraction at each episode, incurring additional inference cost beyond the base agent.
The quality of extracted experience entries and deltas is sensitive to the capability of the underlying LLM, and low-quality extractions may introduce noise into the memory tree.
The retrieval threshold $\tau_{\text{base}}$ (per tree) and the consolidation threshold $K_{\text{cons}}$ are selected per benchmark on small held-out samples, which may not generalize automatically to new task domains without re-tuning.
Additionally, the current framework does not model temporal decay of outdated experience; nodes reflecting stale environment states (\eg{} website layout changes) may persist and mislead future retrieval.

\section*{Ethics Statement}

The data used in this paper comes from publicly available licensed datasets, used for research purposes under their respective licenses. This work complies with the ACL Code of Ethics, and we declare no ethical issues.

\clearpage

\bibliography{custom}

\clearpage

\appendix

\section{Parameter Settings}
\label{app:params}

Table~\ref{tab:params} lists the final threshold and consolidation parameters selected for each benchmark.
For ALFWorld and ScienceWorld, values were determined by inspecting the empirical similarity score distribution on a held-out sample; the configuration used in the main experiment is reported here.
WebShop uses heuristically chosen values, as its product-search embedding space is more homogeneous.

\begin{table}[h]
  \centering
  \small
  \caption{Selected parameter values per benchmark.}
  \label{tab:params}
  \begin{tabular}{lccc}
    \toprule
    \textbf{Benchmark} & $\tau^{\text{task}}_{\text{base}}$ & $\tau^{\text{env}}_{\text{base}}$ & $K_{\text{cons}}$ \\
    \midrule
    ALFWorld    & 0.75 & 0.85 & 5 \\
    ScienceWorld & 0.82 & 0.82 & 3 \\
    WebShop     & 0.85 & 0.85 & 3 \\
    \bottomrule
  \end{tabular}
\end{table}

\section{Memory Token Statistics}
\label{app:token_stats}

Table~\ref{tab:tokens} reports the average number of tokens per stored memory unit for each method, measured on the ALFWorld online evaluation run.
For \sys{}, root and residual nodes are counted separately; all other methods use one entry per memory unit.
Token counts are estimated by whitespace tokenization on the stored text content.
For ReasoningBank, the unit is one episode's distilled lesson set (\ie{}, all items extracted from a single interaction); for AWM, it is one per-task-type workflow file.

\begin{table}[h]
  \centering
  \small
  \caption{Average tokens per memory unit.}
  \label{tab:tokens}
  \begin{tabular}{lc}
    \toprule
    \textbf{Method} & \textbf{Avg.\ tokens / memory} \\
    \midrule
    Synapse            & 448 \\
    AWM                & 59  \\
    ReasoningBank      & 148 \\
    \midrule
    \sys{} (Root)      & 257 \\
    \sys{} (Residual)  & 145 \\
    \bottomrule
  \end{tabular}
\end{table}

\section{Prompt Templates}
\label{app:prompts}

\sys{} uses three categories of prompts: (1)~per-benchmark agent system instructions that define the action space and interaction format; (2)~a memory retrieval header injected between the in-context demonstration and the current task, which varies by method; and (3)~six experience extraction prompts that guide tree updates after each episode.
Experience extraction prompts are instantiated per benchmark; we show the \textbf{ALFWorld} canonical versions below as representative examples.

\subsection{Agent System Instructions}

\begin{promptlisting}{ALFWorld Agent Instruction}
Interact with a household to solve a task. Imagine you are an intelligent
agent in a household environment and your target is to perform actions to
complete the task goal. At the beginning of your interactions, you will be
given the detailed description of the current environment and your goal to
accomplish.
For each of your turn, you will be given the observation of the last turn.
You should choose from two actions: "Thought" or "Action". If you choose
"Thought", you should first think about the current condition and plan for
your future actions, and then output your action in this turn. Your output
must strictly follow this format:"Thought: your thoughts.\n Action: your
next action"; If you choose "Action", you should directly output the action
in this turn. Your output must strictly follow this format:
"Action: your next action".
The available actions are:
1. go to {recep}
2. take {obj} from {recep}
3. put {obj} in/on {recep}
4. open {recep}
5. close {recep}
6. toggle {obj} {recep}
7. clean {obj} with {recep}
8. heat {obj} with {recep}
9. cool {obj} with {recep}
where {obj} and {recep} correspond to objects and receptacles.
After your each turn, the environment will give you immediate feedback.
If the environment outputs "Nothing happened", the previous action is
invalid -- try more options.
Reminder:
1. The action must be chosen from the given available actions.
2. Think when necessary, try to act directly more in the process.
\end{promptlisting}

\begin{promptlisting}{ScienceWorld Agent Instruction}
You are a helpful assistant to do some scientific experiment in an
environment. In the environment, there are several rooms: kitchen,
foundry, workshop, bathroom, outside, living room, bedroom, greenhouse,
art studio, hallway. You should explore the environment and find the items
you need to complete the experiment. You can teleport to any room in one
step. All containers in the environment have already been opened, you can
directly get items from the containers.
For each of your turn, you will be given the observation of the last turn.
You should choose from two actions: "Thought" or "Action". If you choose
"Thought", your output must follow: "Thought: your thoughts.\n Action:
your next action". If you choose "Action": "Action: your next action".
Remember that you can only output one "Action:" per response.

The available actions are:
open OBJ / close OBJ / activate OBJ / deactivate OBJ
connect OBJ to OBJ / disconnect OBJ
use OBJ [on OBJ] / look around / examine OBJ / look at OBJ / read OBJ
move OBJ to OBJ / pick up OBJ / pour OBJ into OBJ / mix OBJ
teleport to LOC / focus on OBJ / wait / wait1
\end{promptlisting}

\begin{promptlisting}{WebShop Agent Instruction}
You are web shopping.
I will give you instructions about what to do.
You have to follow the instructions.
Every round I will give you an observation and a list of available
actions, you have to respond an action based on the state and instruction.
You can use search action if search is available.
You can click one of the buttons in clickables.
An action should be of the following structure:
  search[keywords]
  click[value]
If the action is not valid, perform nothing.
Keywords in search are up to you, but the value in click must be a value
in the list of available actions.
Remember that your keywords in search should be carefully designed.
Your response should use the following format:
  Thought: I think ...
  Action: click[something]
\end{promptlisting}

\subsection{Agent Decision Prompt with Memory Injection}

All methods share the same template structure; only the \texttt{memory\_header} and \texttt{memory\_context} fields differ by method.

\begin{promptlisting}{Agent Decision Prompt Template (memory-injected)}
{instruction}
---
Here is an example for a complete task trajectory.

{examples}
---

{memory_header}

{memory_context}

Now, it's your turn and here is the task.
{task}
\end{promptlisting}

\subsection{DeltaMem Memory Retrieval Headers}

The memory header appears between the ICL demonstration and the current task.
It varies by benchmark to describe the domain-specific contents of each tree.

\begin{promptlisting}{DeltaMem Memory Header -- ALFWorld}
Retrieved from your hierarchical memory system:
* Task Skill Memory -- HOW to solve this task type (Base Skill + Skill
  Deltas as patches)
* Environment Knowledge Memory -- WHERE objects are and HOW to operate
  receptacles/appliances
Note: 'Trigger'/'Applicable scenario' labels describe PAST episodes --
your actual task is at the end.
\end{promptlisting}

\begin{promptlisting}{DeltaMem Memory Header -- ScienceWorld}
Retrieved from your hierarchical memory system:
* Task Skill Memory -- HOW to perform this science task type: exact action
  sequence, which room to find each item, device operation syntax
  (use/activate/connect), and what to watch out for.
* Environment Knowledge Memory -- WHERE specific items are located across
  ScienceWorld rooms, and HOW devices/equipment respond to actions.
Note: 'Trigger'/'Applicable scenario' labels describe PAST episodes --
your actual task is at the end.
\end{promptlisting}

\begin{promptlisting}{DeltaMem Memory Header -- WebShop}
How to use the memory below -- READ CAREFULLY
The memory contains three kinds of entries:
  [Base Skill] -- a worked example trajectory from a SUCCESSFUL past
    shopping episode. This Base Skill ALWAYS applies to your current task.
    Its Trigger label is just an index/category tag, NOT a gating
    condition. You MUST treat it as a second ICL example and imitate its
    reasoning pattern (how it composed search keywords, how it picked the
    candidate from results, the order in which it clicked options, when it
    pressed Buy Now).
  [Skill Delta N] -- a short supplementary patch that adds extra advice ON
    TOP of the Base. Apply a patch only if its Trigger condition literally
    matches your current task; otherwise ignore that patch (but still apply
    the Base).
  [Failure Record / WARN] -- a warning about a trap. NEVER execute its
    procedure; use it only to AVOID the trap it describes.

How to actually use it (mandatory):
  STEP A -- Before your first Action, your Thought MUST contain one
    sentence stating how you are adapting the Base Skill's pattern to
    your current task.
  STEP B -- Write Action: search[...] using YOUR task's product/
    attributes/price, NOT the Base Skill's literal values.
  STEP C -- On the item page, follow the SAME option-selection order
    observed in the Base Skill but click YOUR task's values.
  STEP D -- If any Patch's Trigger literally matches your task, weave its
    advice into Steps B-C. Do NOT blend irrelevant patches.
\end{promptlisting}

\subsection{Task Tree: Experience Extraction Prompts (ALFWorld)}

Four prompts cover the combination of node type (root vs.\ residual) and
episode outcome (success vs.\ failure).
ScienceWorld, WebShop, and Mind2Web use analogous prompts with
domain-specific field descriptions and syntax guidance.

\begin{promptlisting}{Task Tree -- Root Success}
You are a Skill Extractor. Extract a reusable Base Skill from this
successful trajectory.

Environment: {env_description}
Task: {task_description}
Result: SUCCESS ({steps}/{max_steps} steps)

{trajectory}

Output a self-contained Base Skill -- future agents have NO access to
this trajectory:
- `activation_condition`: In plain natural language, describe when this
  skill applies and what distinguishes it from other tasks.
- `execution_procedure`: Concrete step sequence derived from this
  trajectory. Write out each individual action -- do not collapse
  multi-step operations into a single abstract phrase.
- `termination_condition`: When the skill is complete.

Output ONLY the JSON:
{
    "activation_condition": "...",
    "execution_procedure": "...",
    "termination_condition": "..."
}
\end{promptlisting}

\begin{promptlisting}{Task Tree -- Root Failure}
You are a Failure Recorder. This is a FAILURE RECORD -- NOT a skill to
execute. Future agents will see a [WARN] warning with this.

Environment: {env_description}
Task: {task_description}
Result: FAILURE ({steps}/{max_steps} steps)

{trajectory}

- `activation_condition`: In plain natural language, describe what task
  situation this applies to and what wrong assumption caused the failure.
- `execution_procedure`:
  [FAILED]: Actions tried and environment responses showing they failed.
  [UNEXPLORED]: Plausible approaches never attempted.
- `termination_condition`: Leave empty string.

Output ONLY the JSON:
{
    "activation_condition": "...",
    "execution_procedure": "[FAILED]: ...\n[UNEXPLORED]: ...",
    "termination_condition": ""
}
\end{promptlisting}

\begin{promptlisting}{Task Tree -- Residual Success (Node)}
You are a Skill Delta Extractor. Extract ONLY the minimal new patch not
covered by existing skills.

=== EXISTING SKILL MEMORIES ===
{retrieved_task_memory}
=== END ===

Environment: {env_description}
Task: {task_description}
Result: SUCCESS ({steps}/{max_steps} steps)

{trajectory}

Output {"skip": true} ONLY if you can satisfy ALL of the following,
with direct evidence:
1. Identify ONE existing skill whose execution_procedure explicitly lists
   every distinct action type in this trajectory -- quote the exact phrase
   from that skill for each action.
2. No action required a recovery step, different object category, or
   procedural order not covered by that quoted text.
If you cannot quote matching text for even ONE action, you MUST write a
delta.

Otherwise, output the smallest delta (1-3 new observations max):
- `activation_condition`: The specific new condition that makes this delta
  necessary -- must differ from existing triggers.
- `execution_procedure`: New steps only, concrete and self-contained.
- `termination_condition`: When this delta is done.

Output ONLY one of these two JSON formats:
{"skip": true}
OR
{
    "activation_condition": "...",
    "execution_procedure": "...",
    "termination_condition": "..."
}
\end{promptlisting}

\begin{promptlisting}{Task Tree -- Residual Failure (Node)}
You are a Failure Recorder. Record the gap in existing skills that caused
this failure.

=== EXISTING SKILL MEMORIES ===
{retrieved_task_memory}
=== END ===

Environment: {env_description}
Task: {task_description}
Result: FAILURE ({steps}/{max_steps} steps)

{trajectory}

Output {"skip": true} ONLY if an existing failure record describes the
EXACT SAME failure: you must quote the specific failed actions and the
exact environment responses from that record. If the failed action
sequence or environment response differs in any way, you MUST write a new
record.

Otherwise, output the gap as a trap record:
- `activation_condition`: The specific situation that existing skills
  failed to handle.
- `execution_procedure`:
  [FAILED]: What was tried and why it failed.
  [UNEXPLORED]: Approaches never attempted.
- `termination_condition`: Leave empty string.

Output ONLY one of these two JSON formats:
{"skip": true}
OR
{
    "activation_condition": "...",
    "execution_procedure": "[FAILED]: ...\n[UNEXPLORED]: ...",
    "termination_condition": ""
}
\end{promptlisting}

\subsection{Environment Tree: Extraction Prompts (ALFWorld)}

The Environment Tree uses a single prompt template per node type (root or
residual), shared across both success and failure episodes---environmental
observations are valid knowledge regardless of task outcome.

\begin{promptlisting}{Environment Tree -- Root (First Episode)}
You are an Environment Knowledge Extractor. Extract declarative facts
about this household environment from the trajectory -- regardless of
whether the task succeeded or failed, the environmental observations are
valid knowledge.

Environment: {env_description}
Task: {task_description}
Outcome: {result} ({steps}/{max_steps} steps)

{trajectory}

Output self-contained Base Environment Knowledge -- FACTS about the world,
not a procedure:
- `activation_condition`: "Applicable in [environment_type] environments
  where ..." -- key structural features.
- `execution_procedure`: CATEGORY-LEVEL patterns only --
  (A) what TYPES of objects tend to be in what TYPES of locations
    (e.g., "condiment-type objects tend to be on countertops or in kitchen
    drawers"),
  (B) how receptacle/appliance types behave,
  (C) pitfalls observed.
  Write as observations ("X-type objects tend to be on Y"), not commands.
  Do NOT record specific instances (e.g., "cabinet 3 had saltshaker 1") --
  ALFWorld randomizes object placement each episode.
- `termination_condition`: When this knowledge has been fully applied.

Output ONLY the JSON:
{
    "activation_condition": "Applicable in [environment_type] environments
where ...",
    "execution_procedure": "...",
    "termination_condition": "..."
}
\end{promptlisting}

\begin{promptlisting}{Environment Tree -- Residual Node}
You are an Environment Knowledge Extractor. Extract ONLY new environment
facts not covered by existing knowledge -- regardless of whether the task
succeeded or failed.

=== EXISTING ENVIRONMENT KNOWLEDGE ===
{retrieved_env_memory}
=== END ===

Environment: {env_description}
Task: {task_description}
Outcome: {result} ({steps}/{max_steps} steps)

{trajectory}

Output {"skip": true} ONLY when EVERY specific object location AND EVERY
appliance/receptacle interaction rule observed in this trajectory is
already explicitly stated in the existing knowledge above. If even ONE
location or rule is not explicitly covered, you MUST write an update.

Otherwise, output the smallest new update (1-3 new facts max):
- `activation_condition`: "Applicable in [environment_type] environments
  where ..." -- the new structural feature.
- `execution_procedure`: CATEGORY-LEVEL patterns only -- new
  object-type -> location-type tendencies or appliance behavior rules not
  in existing knowledge. Write as observations, not commands. Do NOT record
  specific instances.
- `termination_condition`: When this adaptation is complete.

Output ONLY one of these two JSON formats:
{"skip": true}
OR
{
    "activation_condition": "Applicable in [environment_type] environments
where ...",
    "execution_procedure": "...",
    "termination_condition": "..."
}
\end{promptlisting}

\subsection{Task Tree: Experience Extraction Prompts (ScienceWorld)}

\begin{promptlisting}{Task Tree -- Root Success (ScienceWorld)}
You are a Skill Extractor. Extract a reusable Base Skill from this
successful science experiment trajectory.

Environment: {env_description}
Task: {task_description}
Result: SUCCESS (Steps: {steps}, Reward: {reward}/1.0)

{trajectory}

Output a self-contained Base Skill -- future agents have NO access
to this trajectory:
- `activation_condition`: The science task TYPE and distinguishing
  features (e.g., "boiling tasks where the goal is to heat a liquid
  until it reaches its boiling point, requiring a heat source and
  thermometer").
- `execution_procedure`: Concrete step-by-step procedure using exact
  ScienceWorld syntax (teleport to LOC / pick up OBJ / use OBJ on OBJ
  / focus on OBJ / pour OBJ into OBJ / mix OBJ / activate OBJ /
  connect OBJ to OBJ). Include: (1) which room each item is found in,
  (2) the full action sequence in order, (3) any critical preconditions
  or pitfalls.
- `termination_condition`: When the skill is complete (e.g., reward > 0
  confirmed / target substance reached required state).

Output ONLY the JSON:
{
    "activation_condition": "...",
    "execution_procedure": "...",
    "termination_condition": "..."
}
\end{promptlisting}

\begin{promptlisting}{Task Tree -- Root Failure (ScienceWorld)}
You are a Failure Recorder. This is a FAILURE RECORD -- NOT a skill
to execute. Future agents will see a [WARN] warning with this.

Environment: {env_description}
Task: {task_description}
Result: FAILURE (Steps: {steps}, Reward: {reward}/1.0)
Note: Reward > 0 means some sub-goals were completed before failure.

{trajectory}

- `activation_condition`: Task type + the specific wrong assumption or
  failure condition that caused this outcome.
- `execution_procedure`:
  [FAILED]: Actions attempted and exact environment responses showing
  failure. If reward > 0, identify which sub-goals succeeded before
  breakdown.
  [UNEXPLORED]: Plausible approaches this agent never attempted.
- `termination_condition`: Leave empty string.

Output ONLY the JSON:
{
    "activation_condition": "...",
    "execution_procedure": "[FAILED]: ...\n[UNEXPLORED]: ...",
    "termination_condition": ""
}
\end{promptlisting}

\begin{promptlisting}{Task Tree -- Residual Success (ScienceWorld)}
You are a Skill Delta Extractor. Extract ONLY the minimal new patch
not covered by existing skills.

=== EXISTING TASK SKILL MEMORIES ===
{retrieved_task_memory}
=== END ===

Environment: {env_description}
Task: {task_description}
Result: SUCCESS (Steps: {steps}, Reward: {reward}/1.0)

{trajectory}

Output {"skip": true} ONLY if ALL of the following hold with evidence:
1. Reward is 1.0 (full success).
2. Identify ONE existing skill whose execution_procedure explicitly
   lists every distinct action performed -- quote the exact phrase.
3. No action required a recovery step, different item, or procedural
   order not covered by that quoted text.
If ANY condition fails, you MUST write a delta.

Otherwise, output the smallest delta (1-3 new observations max):
- `activation_condition`: The SPECIFIC NEW condition for this delta.
- `execution_procedure`: NEW steps/rules only, using exact ScienceWorld
  action syntax.
- `termination_condition`: When this delta modification is complete.

Output ONLY one of these two JSON formats:
{"skip": true}
OR
{
    "activation_condition": "...",
    "execution_procedure": "...",
    "termination_condition": "..."
}
\end{promptlisting}

\begin{promptlisting}{Task Tree -- Residual Failure (ScienceWorld)}
You are a Failure Recorder. Record the gap in existing skills that
caused this failure.

=== EXISTING TASK SKILL MEMORIES ===
{retrieved_task_memory}
=== END ===

Environment: {env_description}
Task: {task_description}
Result: FAILURE (Steps: {steps}, Reward: {reward}/1.0)
Note: Reward > 0 means some sub-goals were completed before failure.

{trajectory}

Output {"skip": true} ONLY if an existing failure record describes the
EXACT SAME failure: quote the specific failed actions and exact
environment responses from that record. If any detail differs, write a
new record.

Otherwise, output the gap as a trap record:
- `activation_condition`: The specific situation existing skills failed
  to handle.
- `execution_procedure`:
  [FAILED]: Actions attempted and exact environment responses. Note
  which existing skill guidance was followed but did not work.
  [UNEXPLORED]: Plausible approaches never attempted.
- `termination_condition`: Leave empty string.

Output ONLY one of these two JSON formats:
{"skip": true}
OR
{
    "activation_condition": "...",
    "execution_procedure": "[FAILED]: ...\n[UNEXPLORED]: ...",
    "termination_condition": ""
}
\end{promptlisting}

\subsection{Environment Tree: Extraction Prompts (ScienceWorld)}

Unlike ALFWorld, ScienceWorld rooms have consistent item placement
across episodes, so the Env-Tree records specific room-to-item
mappings rather than category-level tendencies.

\begin{promptlisting}{Environment Tree -- Root (ScienceWorld)}
You are an Environment Knowledge Extractor. Extract declarative facts
about this ScienceWorld environment -- regardless of task outcome,
the observations are valid knowledge.

Environment: {env_description}
Task: {task_description}
Outcome: Steps={steps}, Reward={reward}/1.0

{trajectory}

Important: ScienceWorld rooms have CONSISTENT item placement across
episodes. Record SPECIFIC room->item mappings (e.g., "the thermometer
is in the kitchen"), not just category-level tendencies.

- `activation_condition`: "Applicable in ScienceWorld environments
  where ..." -- describe the task context or room configuration.
- `execution_procedure`: Factual observations:
  (A) specific item locations by room (e.g., "thermometer: kitchen",
    "battery: workshop"),
  (B) device/equipment operation rules with exact action syntax
    (e.g., "use thermometer on OBJ to read temperature"),
  (C) efficient room search order for common items,
  (D) any pitfalls observed.
  Write as facts, not commands.
- `termination_condition`: When this environment knowledge has been
  fully applied.

Output ONLY the JSON:
{
    "activation_condition": "Applicable in ScienceWorld environments
where ...",
    "execution_procedure": "...",
    "termination_condition": "..."
}
\end{promptlisting}

\begin{promptlisting}{Environment Tree -- Residual Node (ScienceWorld)}
You are an Environment Knowledge Extractor. Extract ONLY new
environment facts not covered by existing knowledge -- regardless of
task outcome.

=== EXISTING ENVIRONMENT KNOWLEDGE ===
{retrieved_env_memory}
=== END ===

Environment: {env_description}
Task: {task_description}
Outcome: Steps={steps}, Reward={reward}/1.0

Important: ScienceWorld rooms have CONSISTENT item placement. Record
specific room->item mappings when newly observed.

Output {"skip": true} ONLY when EVERY item location AND EVERY
device/equipment interaction rule observed in this trajectory is
already explicitly stated in the existing knowledge above.

Otherwise, output the smallest new update (1-3 new facts max):
- `activation_condition`: "Applicable in ScienceWorld environments
  where ..." -- the new context covered.
- `execution_procedure`: NEW facts only -- specific item locations or
  device rules not in existing knowledge. Written as facts, not
  commands.
- `termination_condition`: When this environment adaptation is
  complete.

Output ONLY one of these two JSON formats:
{"skip": true}
OR
{
    "activation_condition": "Applicable in ScienceWorld environments
where ...",
    "execution_procedure": "...",
    "termination_condition": "..."
}
\end{promptlisting}

\subsection{Task Tree: Experience Extraction Prompts (WebShop)}

\begin{promptlisting}{Task Tree -- Root Success (WebShop)}
You are saving a SUCCESSFUL WebShop trajectory as a reference example
for future agents -- like a curated 1-shot ICL exemplar, NOT an
abstract instruction manual.

Product Category: {env_description}
Task: {task_description}
Result: SUCCESS ({steps}/{max_steps} steps)

=== RAW TRAJECTORY ===
{trajectory}
=== END ===

Clean and condense the raw trajectory into a SHORT illustrative dialog
the future agent can IMITATE. Keep Observation -> Thought -> Action
format.

Rules:
- KEEP every Action that contributed to success (search, candidate
  click, option selections, Buy Now). DROP failed/backtracked attempts.
- For each kept Observation, TRIM noise but preserve lines that justify
  the next Action. One Observation block <= 60 words.
- Add a brief Thought before each Action explaining WHY (<= 25 words).
- End after the successful Action: click[Buy Now].

- `activation_condition`: Short noun phrase naming when this exemplar
  is relevant. Specific to the product category but not exact
  constraints. <= 30 words.
- `execution_procedure`: The cleaned trajectory text. Multi-line,
  plain text, Obs/Thought/Action format.
- `termination_condition`: "After Action: click[Buy Now] succeeds with
  full reward."

Output ONLY the JSON (escape newlines as \n in execution_procedure):
{
    "activation_condition": "...",
    "execution_procedure": "Observation: ...\nThought: ...\nAction:
search[...]\n...\nAction: click[Buy Now]",
    "termination_condition": "..."
}
\end{promptlisting}

\begin{promptlisting}{Task Tree -- Root Failure (WebShop)}
You are a Failure Recorder. This is a FAILURE RECORD -- NOT a skill
to execute. Future agents will see a [WARN] warning with this.

Product Category: {env_description}
Task: {task_description}
Result: FAILURE ({steps}/{max_steps} steps)

{trajectory}

- `activation_condition`: Describe what WebShop task situation this
  applies to and what wrong assumption caused the failure (e.g.,
  too-broad search keywords, picking without checking constraints,
  missing option selection before Buy Now).
- `execution_procedure`:
  [FAILED]: Actions tried and how WebShop responded (e.g., "search
  returned 0 results", "Buy Now pressed without selecting size").
  [UNEXPLORED]: Plausible alternative keywords / candidates / options
  never attempted.
- `termination_condition`: Leave empty string.

Output ONLY the JSON:
{
    "activation_condition": "...",
    "execution_procedure": "[FAILED]: ...\n[UNEXPLORED]: ...",
    "termination_condition": ""
}
\end{promptlisting}

\begin{promptlisting}{Task Tree -- Residual Success (WebShop)}
You are writing a Skill Patch that supplements an existing Base Skill
-- NOT another standalone procedure.

=== EXISTING SKILL CHAIN (Base + previous patches) ===
{retrieved_task_memory}
=== END ===

Product Category: {env_description}
Task: {task_description}
Result: SUCCESS ({steps}/{max_steps} steps)

{trajectory}

Write a patch ONLY if the trajectory reveals a new tactic or
special-case observation the base skill does NOT already cover.
Examples: a new search-keyword trick for this sub-category, a new
option-selection ordering, a common failure mode and how to avoid it.

Rules:
- DO NOT restate the base procedure.
- Write ONLY what is different/additional for this sub-category.
- 1-4 short bullets, <= 80 words in execution_procedure.
- activation_condition names the SPECIFIC trigger.
- execution_procedure starts with "In addition to the base:" or
  "Replace base step N with:".

Output {"skip": true} if the trajectory shows nothing new beyond the
base skill.

Output ONLY one of these two JSON formats:
{"skip": true}
OR
{
    "activation_condition": "When [specific sub-category / constraint
combo] applies...",
    "execution_procedure": "In addition to the base: ...",
    "termination_condition": ""
}
\end{promptlisting}

\begin{promptlisting}{Task Tree -- Residual Failure (WebShop)}
You are recording a Failure Patch -- a warning supplementing the
existing Base Skill, not a standalone failure log.

=== EXISTING SKILL CHAIN (Base + previous patches/warnings) ===
{retrieved_task_memory}
=== END ===

Product Category: {env_description}
Task: {task_description}
Result: FAILURE ({steps}/{max_steps} steps)

{trajectory}

Identify the SPECIFIC trap this trajectory hit and write a SHORT
warning to help future agents avoid it.

Rules:
- DO NOT restate base procedure. Just the trap.
- activation_condition: name the specific sub-category + symptom.
- execution_procedure (<= 80 words): start with [TRAP]: followed by
  what caused the wrong action, then [FIX]: followed by the corrective
  tactic.
- termination_condition: leave empty.

Output {"skip": true} ONLY if an existing failure record already warns
about the exact same trap.

Output ONLY one of these two JSON formats:
{"skip": true}
OR
{
    "activation_condition": "When [specific sub-category + trap
symptom]...",
    "execution_procedure": "[TRAP]: ...\n[FIX]: ...",
    "termination_condition": ""
}
\end{promptlisting}

\subsection{Environment Tree: Extraction Prompts (WebShop)}

In WebShop, the Env-Tree captures UI behavior patterns for each
product category rather than physical scene layouts.

\begin{promptlisting}{Environment Tree -- Root (WebShop)}
You are a WebShop Site Knowledge Extractor. Extract declarative FACTS
about WebShop's UI behaviour for this product category, written so
they remain useful when a future agent shops in the same category for
a different item.

Product Category: {env_description}
Task this episode: {task_description}
Outcome: {result} ({steps}/{max_steps} steps)

{trajectory}

Output self-contained Base Site Knowledge -- FACTS, NOT step-by-step
procedure:
- `activation_condition`: "Applicable when shopping for
  {env_description} on WebShop." Add structural traits observed (e.g.,
  "items in this category expose size and color as required options
  before Buy Now").
- `execution_procedure`: Three groups of FACTS:
  (A) Option dimensions exposed for this category (size? color? scent?
    pack-size?) and which are REQUIRED before Buy Now for full reward.
  (B) Search-keyword patterns that worked -- as patterns, not literal
    strings.
  (C) Pitfalls observed (e.g., "Buy Now without selecting size ->
    partial reward only").
- `termination_condition`: When the agent has applied these facts to
  inform a correct selection.

Output ONLY the JSON:
{
    "activation_condition": "Applicable when shopping for
{env_description} on WebShop ...",
    "execution_procedure": "(A) ... (B) ... (C) ...",
    "termination_condition": "..."
}
\end{promptlisting}

\begin{promptlisting}{Environment Tree -- Residual Node (WebShop)}
You are writing a Site Knowledge Patch -- a small note supplementing
existing site knowledge, NOT a restatement of it.

=== EXISTING SITE KNOWLEDGE ===
{retrieved_env_memory}
=== END ===

Product Category: {env_description}
Task this episode: {task_description}
Outcome: {result} ({steps}/{max_steps} steps)

{trajectory}

Note a NEW site fact this trajectory revealed that the existing
knowledge does not already state (e.g., a required option dimension,
a pagination detail, a quantity default).

Rules:
- DO NOT restate existing knowledge. Just the new fact.
- 1-3 short bullets, <= 60 words in execution_procedure.
- activation_condition names the SPECIFIC sub-category trigger.

Output {"skip": true} ONLY if every UI fact observed here is already
in the existing knowledge above.

Output ONLY one of these two JSON formats:
{"skip": true}
OR
{
    "activation_condition": "When shopping for {env_description} on
WebShop ...",
    "execution_procedure": "...",
    "termination_condition": ""
}
\end{promptlisting}

\end{document}